\documentclass[letterpaper]{article} 
\usepackage{aaai2026}  
\usepackage{times}  
\usepackage{helvet}  
\usepackage{courier}  
\usepackage[hyphens]{url}  
\usepackage{graphicx} 
\usepackage{natbib}  
\usepackage{caption} 
\usepackage{amssymb}
\usepackage{amsmath}
\usepackage{amssymb}
\usepackage{amsmath}
\usepackage{amsthm}
\usepackage{bm}
\usepackage{booktabs}
\usepackage{graphicx}
\usepackage{caption}
\usepackage{enumerate}
\usepackage{comment}
\newcommand{\vcdim}{\operatorname{VCdim}}
\newcommand{\pdim}{\operatorname{Pdim}}
\newcommand{\sgn}{\operatorname{sgn}}
\newcommand{\real}{\mathbb{R}}
\theoremstyle{definition}

\newtheorem{assumption}{Assumption}
\theoremstyle{plain}
\newtheorem{theorem}{Theorem}
\newtheorem{proposition}{Proposition}
\newtheorem{fact}{Fact}
\newtheorem{lemma}{Lemma}
\theoremstyle{remark}
\newtheorem{remark}{Remark}
\usepackage[utf8]{inputenc}

\usepackage{algorithm}
\usepackage{algorithmic}

\usepackage{newfloat}
\usepackage{listings}
\DeclareCaptionStyle{ruled}{labelfont=normalfont,labelsep=colon,strut=off} 
\floatstyle{ruled}
\newfloat{listing}{tb}{lst}{}
\floatname{listing}{Listing}
\title{Minimum-Cost Network Flow with Dual Predictions}
\author{
    Zhiyang Chen\textsuperscript{\rm 1}, Hailong Yao\textsuperscript{\rm 2,3}\thanks{Corresponding author (E-mail: hailongyao@ustb.edu.cn).}, Xia Yin\textsuperscript{\rm 1}\\
}
\affiliations{
    \textsuperscript{\rm 1}Tsinghua University\\
    \textsuperscript{\rm 2}University of Science and Technology Beijing\\
    \textsuperscript{\rm 3}Key Laboratory of Advanced Materials and
Devices for Post-Moore Chips, Ministry of Education of China

}

\usepackage{bibentry}

\begin{document}

\maketitle

\begin{abstract}
Recent work has shown that machine-learned predictions can provably improve the performance of classic algorithms. In this work, we propose the first minimum-cost network flow algorithm augmented with a dual prediction. Our method is based on a classic minimum-cost flow algorithm, namely $\varepsilon$-relaxation. We provide time complexity bounds in terms of the infinity norm prediction error, which is both consistent and robust. We also prove sample complexity bounds for PAC-learning the prediction. We empirically validate our theoretical results on two applications of minimum-cost flow, i.e., traffic networks and chip escape routing, in which we learn a fixed prediction, and a feature-based neural network model to infer the prediction, respectively. Experimental results illustrate $12.74\times$ and $1.64\times$ average speedup on two applications.
\end{abstract}

\section{Introduction}
Minimum-cost network flow is an important computational problem in combinatorial optimization, with various applications in computer science and operations research, including design automation~\cite{yan10recent}, computer vision~\cite{wang2019mussp}, and scheduling~\cite{ahuja1995applications}. Moreover, many network optimization problems, e.g., disjoint paths, maximum flow, minimum cut, matching, and assignment, are special cases of minimum-cost flow. Therefore, fast minimum-cost flow algorithms are essential for efficiently solving many practical optimization problems.

A bunch of minimum-cost flow algorithms have been developed throughout the last 40 years~\cite{bertsekas1998network,ahuja93network}, but the minimum-cost flow problem is far from being fully solved. On the one hand, although minimum-cost flow can be solved with polynomial time complexity, it is time-consuming in practice to solve a large-scale minimum-cost flow instance, e.g., a network with millions of nodes and edges. Academics are still actively exploring practically fast network flow solvers~\cite{kara22parallel}. On the other hand, the theoretical analysis of existing minimum-cost flow algorithms is not compatible with the algorithm behaviors in practice. Traditional worst-case time complexity analysis is often overly pessimistic, yielding loose complexity bounds. For instance, although the successive shortest path algorithm may theoretically require an exponential number of iterations to find the optimum~\cite{zadeh73bad}, it usually runs relatively faster on minimum-cost flow instances in practice.

Recent work has shown that leveraging machine-learned predictions can improve the worst-case performance of classic algorithms~\cite{alps}, namely {\em algorithms with predictions} or {\em learning-augmented algorithms}. It is a beyond-worst-case framework for algorithm design and analysis. In this framework, we assume the algorithm is given a black-box prediction which outputs information about the problem instance. We can leverage this prediction to improve the performance of the algorithm. In practice, such predictions are obtained by machine learning techniques and are hence imperfect. Therefore, we hope the algorithm is both {\em consistent} and {\em robust}: It works excellent if the prediction error is small and preserves the classic worst-case performance bound if the prediction is completely wrong.

Previous works study algorithms with predictions for many network optimization problems, including matching~\cite{dinitz21faster}, maximum flow~\cite{davies23predictive}, and global minimum cut~\cite{niaparast25faster}. Therefore, it is natural to ask the question: {\em Can we use learning-based predictions to accelerate minimum-cost network flow algorithms?}

In this work, we propose the first minimum-cost flow algorithm with dual predictions, based on a classic algorithm, $\varepsilon$-relaxation~\cite{bertsekas86distributed}. We show that a dual prediction can provably and practically accelerate $\varepsilon$-relaxation. One may wonder why we study $\varepsilon$-relaxation instead of other more popular minimum-cost algorithms, such as network simplex. We will discuss the advantages of $\varepsilon$-relaxation in preliminaries.

\subsection{Our contributions}\label{sec:contribution}
\noindent\textbf{Theoretical results}. On the theoretical side, we show that the complexity of the $\varepsilon$-relaxation algorithm for minimum-cost flow can be provably improved by a dual prediction.

For a network instance with $n$ nodes and $m$ edges, let $p^*$ be an arbitrary optimal dual solution, and $\hat{p}$ be the predicted solution, we show that warm-starting $\varepsilon$-relaxation with preprocessing solves minimum-cost flow in $O\left(\min\left\{n^3\|\hat{p}-p^*\|_{\infty},\,n^4C\right\}\right)$ time, where $C$ is the maximum of edge costs (Theorem~\ref{thm:time-complexity}). Combining this algorithm with a cost-scaling trick, we can improve this bound to $O\left(\min\left\{n^3\log\|\hat{p}-p^*\|_{\infty},n^3\log(nC)\right\}\right)$ (Theorem~\ref{thm:cost-scaling}). This bound reduces to the $O(n^3\log(nC))$ worst-case complexity of $\varepsilon$-relaxation if the prediction is completely erroneous, and efficiently reduces the running time if the prediction error is small. Therefore, our algorithm is both consistent and robust. Specially, for 0/1 flow instances\footnote{A minimum-cost flow instance is 0/1 flow if the capacities satisfy $b_{ij}=0$ and $c_{ij}=1$ for each $(v_i,v_j)\in E$. A natural example is the minimum-cost bipartite matching.}, the time bound can be improved to $O\left(\min\left\{mn\log\|\hat{p}-p^*\|_{\infty},mn\log(nC)\right\}\right)$.

We also show that we can PAC-learn the prediction from data with a small sample complexity, using the framework of data-driven algorithm design (Theorems~\ref{thm:generalization} and~\ref{thm:nn}).

\noindent\textbf{Empirical results}. On the empirical side, we evaluate our algorithm on two applications of minimum-cost flow: traffic networks and escape routing. For traffic networks, we can learn a fixed prediction if the network topology is fixed and the edge costs are random variables. For escape routing, we can learn a feature-based predictor using convolutional neural networks (CNN) based on labeled data of routing instances. Experimental results show that we achieve about $6.2\sim 21.4\times$ acceleration on real-world traffic networks, and about $1.1\sim 2.3\times$ acceleration for escape routing.

\subsection{Related work}
\noindent\textbf{Minimum-cost flow algorithms}. Network flow has been well studied. We refer readers to, e.g., \citet{bertsekas1998network} and~\citet{ahuja93network} for a comprehensive survey. The simplest minimum-cost flow algorithm is successive shortest path (SSP), which is an extension of Ford-Fulkerson for maximum flow. However, the complexity of SSP grows linearly in terms of the cost range. Subsequent works propose polynomial-time algorithms for minimum-cost flow, including capacity scaling~\cite{edmonds72theoretical}, cost scaling~\cite{goldberg87solving}, and cycle-canceling~\cite{Klein66primal}. The $\varepsilon$-relaxation~\cite{bertsekas86distributed} is a kind of cost scaling algorithm.

The most popular and practically efficient algorithm is network simplex (NS), which is a variant of the simplex method specialized for network flow. \citet{orlin97polynomial} shows that a special type of NS solves minimum-cost flow in polynomial time. However, the time complexity of the general version of NS implemented in practice remains unclear.

The state-of-the-art complexity is due to \citet{brand23deterministic}, which presents an almost-linear time algorithm. However, this approach is purely theoretical.

\noindent\textbf{Algorithms with predictions}. \citet{kraska18case} first propose incorporating machine-learned predictions into data structures. This area has then become popular in various fields. We refer readers to the website~\cite{alps} for a full publication list.

For network optimization with predictions, \citet{dinitz21faster} propose a weighted bipartite matching algorithm with dual predictions. \citet{chen22faster} improve this result and extend it to other graph problems. \citet{davies23predictive} propose a maximum flow algorithm with primal predictions.

\citet{chen22faster} obtains an $O(m^{3/2}n+(mn+m\log m)\|\hat{p}-p^*\|_0)$ algorithm for the special case, min-cost 0/1 flow. Their approach is based on a reduction from bipartite matching and is hence not practical. Their bound is not better than our $O(mn\log\|\hat{p}-p^*\|_\infty)$ bound for 0/1 flow. Also, their bound is based on 0-norm error, which is the number of indices such that the prediction is different from the optimal solution. In practice, it is hard for a learned model to predict exact values for each component. Therefore, our bound is more practical.

\citet{sakaue22discrete} studies discrete energy minimization for computer vision using tools from discrete convex analysis. The dual of min-cost flow can be seen as a special case of this problem. They provide an $O(mn^2\|\hat{p}-\hat{p}^*\|_\infty)$ algorithm, which is worse than our $O(n^3\log\|\hat{p}-\hat{p}^*\|_\infty)$ bound, both in terms of prediction error and graph size. We also emphasize that our min-cost flow algorithm is practical. Both \citet{chen22faster} and \citet{sakaue22discrete} study purely theoretical algorithms.

\noindent\textbf{Data-driven algorithm design}. Another line of research that combines algorithm design and machine learning techniques studies the sample complexity of learning an algorithm configuration from a class of parameterized algorithms. \citet{gupta17pac} propose the framework of data-driven algorithm design. Subsequent work provides sample complexity guarantees for a variety of parameterized algorithms~\cite{blum21learning,balcan21how,balcan22structural,balcan24learning,cheng24sample,chen25learning}. In this work, we will analyze the sample complexity of learning the dual prediction for $\varepsilon$-relaxation, following the philosophy of data-driven algorithm design.


\section{Preliminaries}\label{sec:preliminaries}
\subsection{Linear minimum-cost network flow}
\noindent\textbf{Problem formulation}. Let $G=(V,E)$ be a directed graph with $|V|=n$ nodes and $|E|=m$ edges. Each edge $(v_i,v_j)$ is associated with a cost $a_{ij}$ and a pair of capacity $(b_{ij},c_{ij})$ such that $0\le b_{ij}\le c_{ij}$. Each node $v_i$ has a supply $s_i\in\real$. (If $s_i<0$, it represents $v_i$ has a demand $-s_i$.) The {\em linear minimum-cost network flow problem} (MCF) is formulated as the following:
\begin{align*}
    \min_x &\sum_{(v_i,v_j)\in E}a_{ij}x_{ij},\\
    \text{subj. to} & \sum_{(v_i,v_j)\in E}x_{ij}-\sum_{(v_j,v_i)\in E}x_{ji}=s_i,\ \forall\,v_i\in V,\\
    & b_{ij}\le x_{ij}\le c_{ij},\ \forall\,(v_i,v_j)\in E.
\end{align*}
We also make the following assumption:
\begin{assumption}
    All capacities and supplies are integers so that the optimal solution is integral. Let $C=\max_{(v_i,v_j)\in E}|a_{ij}|$ be the maximum scale of edge costs.
\end{assumption}

\noindent\textbf{Duality}. The dual problem of MCF is \begin{equation*}
    \max_p \sum_{(v_i,v_j)\in E}q_{ij}(p_i-p_j)+\sum_{v_i\in V}s_ip_i,
\end{equation*} where \begin{align*}
    q_{ij}(p_i-p_j)=\left\{\begin{array}{ll}
         (a_{ij}+p_j-p_i)b_{ij} & \text{if}\ a_{ij}+p_j-p_i\ge 0, \\
         (a_{ij}+p_j-p_i)c_{ij} & \text{if}\ a_{ij}+p_j-p_i<0.
    \end{array}\right.
\end{align*} Therefore, a pair of primal solution $x$ and dual solution $p$ is optimal if $x$ is feasible and $(x,p)$ satisfies {\em complementary slackness} (CS), i.e., $\forall\,(v_i,v_j)\in E$, \begin{equation}
    \begin{aligned}
        &x_{ij}<c_{ij}\ \Rightarrow p_i-p_j-a_{ij}\le 0,\\
        &x_{ij}>b_{ij}\ \Rightarrow p_i-p_j-a_{ij}\ge 0.
    \end{aligned}\label{eqn:cs}
\end{equation}
\subsection{The $\varepsilon$-relaxation algorithm}
The $\varepsilon$-relaxation~\cite{bertsekas86distributed} is a dual algorithm for the linear network flow problem. The algorithm considers a relaxed CS condition (\ref{eqn:cs}), namely $\varepsilon$-CS: For each edge $(v_i,v_j)$, we have \begin{equation}
    \begin{aligned}
    &x_{ij}<c_{ij}\ \Rightarrow p_i-p_j-a_{ij}\le \varepsilon,\\
    &x_{ij}>b_{ij}\ \Rightarrow p_i-p_j-a_{ij}\ge -\varepsilon.
    \end{aligned}
\end{equation} We call $p_i-p_j-a_{ij}$ the {\em reduced cost} of edge $(v_i,v_j)$. The algorithm maintains an $\varepsilon$-CS solution pair $(x,p)$, and iteratively reduce the surplus \begin{equation*}
    g_i=\sum_{(v_j,v_i)\in E}x_{ji}-\sum_{(v_i,v_j)\in E}x_{ij}+s_i
\end{equation*} of each node $v_i$ to zero to make $x$ a feasible flow. The algorithmic detail of $\varepsilon$-relaxation is introduced in the appendix.

It can be shown that the algorithm outputs an optimal solution if $\varepsilon$ is small enough~\cite{bertsekas86distributed}. A pessimistic optimality condition is $\varepsilon<\frac{1}{n}$, but we can relax the bound if the graph satisfies specific structures. The time complexity of $\varepsilon$-relaxation is $O(n^3\varepsilon^{-1}C)$, which is $O(n^4C)$ by taking $\varepsilon=\frac{1}{n+1}$. Applying the classic cost-scaling trick improves this bound to $O(n^3\log(nC))$.

It is advantageous to consider $\varepsilon$-relaxation with predictions, compared with other minimum-cost flow algorithms, such as network simplex:
\begin{itemize}
    \item The $\varepsilon$-relaxation is a dual algorithm, making it easier to learn a prediction. For a network with $n$ nodes and $m$ edges, the prediction is required to predict only $n$ (resp. $m=O(n^2)$) solution variables for a dual (resp. primal) algorithm.
    \item The $\varepsilon$-relaxation is easy to parallelize~\cite{parallel97beraldi}. Many serial minimum-cost flow algorithms, including network simplex, are hard to massively parallelize. The iterations of nodes in $\varepsilon$-relaxation are decoupled, making it suitable for parallel acceleration~\cite{lin20abcdplace}.
    \item The $\varepsilon$-relaxation is mathematically easy to analyze. Although network simplex is state-of-the-art in practice, its theoretical time complexity remains unclear.
\end{itemize}

\section{Dual Prediction for $\varepsilon$-Relaxation}\label{sec:algo}
In this section, we introduce and analyze the $\varepsilon$-relaxation with a dual prediction. We assume the algorithm is given a predicted dual solution $\hat{p}$ and use $p^*$ to denote an arbitrary optimal dual solution. Proofs of technical lemmas are omitted to the appendix.

\noindent\textbf{Notations}. For a path $H$ of $G=(V,E)$, it may contain both forward and backward edges of $G$, following the convention of network flow\footnote{An edge $(u,v)$ is a forward (resp. backward) edge of $G=(V,E)$ if $(u,v)\in E$ (resp. $(v,u)\in E$).}. We use $H^+$ and $H^-$ to denote the set of forward and backward edges of $H$, respectively. We also use $s(H)$ and $t(H)$ to denote the start and end nodes of $H$. We say a simple path is {\em unblocked with respect to} $x$, if $x_{ij}<c_{ij}$ for any $(v_i,v_j)\in H^+$ and $x_{ij}>b_{ij}$ for any $(v_i,v_j)\in H^-$. In other words, a path $H$ is unblocked if we can send positive flow along $H$ to $x$ from $s(H)$ to $t(H)$ without violating the capacities.

\begin{algorithm}[htb]
\caption{Warm-start $\varepsilon$-relaxation.}
\label{alg:warm-start}
\textbf{Input}: A graph $G=(V,E)$ and a dual prediction $\hat{p}$.\\
\textbf{Output}: The optimal flow of $G$.
\begin{algorithmic}[1] 
\STATE Let $\hat{p}_{\min}\leftarrow\min_{1\le i\le n}\hat{p}_i$;
\STATE Let $\hat{p}\leftarrow\min\left\{\hat{p}-\hat{p}_{\min},(n-1)C\right\}$. (Clip the values of $\hat{p}$ to make $0\le\hat{p}\le (n-1)C$.)
\STATE Run $\varepsilon$-relaxation with the initial dual solution $\hat{p}$;
\end{algorithmic} 
\end{algorithm}
\subsection{Vanilla warm-start $\varepsilon$-relaxation}
We first preprocess the predicted dual and run $\varepsilon$-relaxation with the initial dual solution $\hat{p}$. See Algorithm~\ref{alg:warm-start} for a description of our algorithm. Note that we have the following fact, since the dual cost only depends on $p_i-p_j$ for each pair of $(v_i,v_j)\in E$. Therefore, for simplicity, we first shift $\hat{p}$ to make $\min_i p_i=0$.
\begin{fact}\label{fact:shift}
    The dual problem of network flow is shift-invariant. If $p^*$ is an optimal solution, $p^*+c$ for any constant $c$ is also an optimal solution.
\end{fact}
We also have the following lemma. Therefore, after preprocessing, the prediction error of the initial dual solution will not increase.
\begin{lemma}\label{lmm:preprocessing}
    There exists an optimal dual solution $p^*$ such that $0\le p^*\le (n-1)C$.
\end{lemma}
\begin{proof}
By Fact 1, there exists an optimal solution such that $\min_i p^*_i=0$. It suffices to show $p^*\le (n-1)C$.

Proof by contradiction. Suppose for any optimal solution $p^*\ge 0$ with $p^*_s=0$ for $v_s\in V$, there exists $v_t\in V$ such that $p^*_t>(n-1)C$. We can always find an $v_s-v_t$ cut that partitions $V$ into two disjoint subsets, $S$ and $T$, which satisfies $v_s\in S$, $v_t\in T$ and for any $v_i\in S$, $v_j\in T$, $|p^*_i-p^*_j|>C$. (Otherwise, there exists a path $H$ from $v_s$ to $v_t$, such that each edge $(v_i,v_j)\in H$ satisfies $|p^*_i-p^*_j|\le C$, and thus $p^*_t\le p^*_s+|H|C\le (n-1)C$.) Since $p^*_s<p^*_t$, the dual values $p^*$ of all nodes in $S$ are smaller than those in $T$.

Since $p^*$ is dual optimal, there exists a flow $x^*$ such that $(x^*,p^*)$ satisfies complementary slackness. For each forward or backward edge $(v_i,v_j)$ of $G$ in the $v_s-v_t$ cut, we have $p^*_j-p^*_i>C$. Therefore, we can decrease the dual values $p^*$ of each node in $T$ until a cut edge $(v_i,v_j)$ satisfies $p^*_j-p^*_i=C$ without violating complementary slackness, and hence retain optimality. This is a contradiction.
\end{proof}
\begin{theorem}\label{thm:time-complexity}
    The time complexity of Algorithm~\ref{alg:warm-start} is $O\left(\min\left\{n^3+n^2\varepsilon^{-1}\|\hat{p}-p^*\|_{\infty},\,n^3\varepsilon^{-1}C\right\}\right)$. Specially, if the input is a 0/1 flow instance, the bound can be improved to $O\left(\min\left\{mn+m\varepsilon^{-1}\|\hat{p}-p^*\|_{\infty},\,mn\varepsilon^{-1}C\right\}\right)$.
\end{theorem}
\begin{proof}
    By Lemma~\ref{lmm:preprocessing}, after preprocessing, the infinity norm prediction error is at most $(n-1)C$, it suffices to prove the $O(n^3+n^2\varepsilon^{-1}\|\hat{p}-p^*\|_{\infty})$ bound (and $O(mn+m\varepsilon^{-1}\|\hat{p}-p^*\|_{\infty})$ for 0/1 flow).
    
    Following the original analysis of $\varepsilon$-relaxation, we first define a few notations. For any path $H$ in $G$, we define the {\em reduced cost length} of $H$ is the sum of each edge's reduced cost, \begin{align*}
        &d_H(p)=\max\{0,\,\sum_{(v_i,v_j)\in H^+}(p_i-p_j-a_{ij})\\
        &\qquad\qquad-\sum_{(v_i,v_j)\in H^-}(p_i-p_j-a_{ij})\}\\
        =&\max\{0,\,p_{s(H)}-p_{t(H)}-\sum_{(v_i,v_j)\in H^+}a_{ij}+\sum_{(v_i,v_j)\in H^-}a_{ij}\}.
    \end{align*} For any dual solution $p$ and primal solution $x$ satisfying the flow conservation and capacity constraints (i.e., $x$ is a feasible flow), we let \begin{equation*}
        D(p,x)=\max_H\{d_H(p)\ |\ H\text{ is a simple unblocked path w.r.t. }x\}.
    \end{equation*} If no simple unblocked path exists, we simply let $D(p,x)=0$. We let \begin{equation*}
        \beta(p)=\min_x\left\{D(p,x)\ |\ x\text{ is a feasible flow}\right\}.
    \end{equation*}
    We can bound $\beta(\hat{p})$ with the infinity norm prediction error.
    \begin{lemma}\label{lmm:betaupperbound}
        For warm-start dual solution $\hat{p}$, we have $\beta(\hat{p})\le 2\|\hat{p}-p^*\|_\infty$.
    \end{lemma}
    It is shown that the time complexity of $\varepsilon$-relaxation is upper bounded by $O(n^3+n^2\beta(p_0)/\varepsilon)$ for an initial solution $p_0$. If the input is a 0/1 flow instance, the time complexity is $O(mn+m\beta(p_0)/\varepsilon)$. (See Proposition 4.1 of Chapter 5, \citet{bertsekas89parallel}). Therefore, our bound follows directly from Lemma~\ref{lmm:betaupperbound}.

    Now, it suffices to prove Lemma~\ref{lmm:betaupperbound}. We have \begin{align}
        \beta(\hat{p})=&\min_{x} D(\hat{p},x)\nonumber\\
        \le& \,D(\hat{p},x^*)\nonumber\\
        =&\max_H\left\{d_H(\hat{p})\ |\ H\text{ is unblocked in }x^*\right\}\nonumber\\
        =&\max_H\{0,\,\hat{p}_{s(H)}-\hat{p}_{t(H)}-\sum_{(v_i,v_j)\in H^+}a_{ij}\nonumber\\
        &\qquad+\sum_{(v_i,v_j)\in H^-}a_{ij}\ \Bigg|\ H\text{ is unblocked in }x^*\}.\label{agn:inftybound}
    \end{align}
    Note that for any unblocked path $H$ in the optimal flow $x^*$, we have \begin{equation*}
        p^*_{s(H)}-p^*_{t(H)}-\sum_{(v_i,v_j)\in H^+}a_{ij}+\sum_{(v_i,v_j)\in H^-}a_{ij}\le 0.
    \end{equation*} This follows from the fact that the optimal solution $(x^*,p^*)$ satisfies complementary slackness so that the reduced cost $p^*_i-p^*_j-a_{ij}$ of each edge $(v_i,v_j)$ in $H$ must be non-positive.

    Therefore, for any dual solution $p$, we have \begin{align*}
        &\,p_{s(H)}-p_{t(H)}-\sum_{(v_i,v_j)\in H^+}a_{ij}+\sum_{(v_i,v_j)\in H^-}a_{ij}\\
        =&\left(p^*_{s(H)}-p^*_{t(H)}-\sum_{(v_i,v_j)\in H^+}a_{ij}+\sum_{(v_i,v_j)\in H^-}a_{ij}\right)\\
        &\quad+(p_{s(H)}-p_{t(H)})-(p^*_{s(H)}-p^*_{t(H)})\\
        \le&\,(p_{s(H)}-p_{t(H)})-(p^*_{s(H)}-p^*_{t(H)})\\
        \le&\,\left|p_{s(H)}-p^*_{s(H)}\right|+\left|p_{t(H)}-p^*_{t(H)}\right|\\
        \le&\,2\|p-p^*\|_{\infty}.
    \end{align*}
    Plugging this bound into (\ref{agn:inftybound}) yields the desired result $\beta(\hat{p})\le 2\|\hat{p}-p^*\|_\infty$.
\end{proof}
We obtain a time bound $O(\min\left\{n^3\|\hat{p}-p^*\|_{\infty},\,n^4C\right\})$, if we set $\varepsilon=\frac{1}{n+1}$ to ensure optimality.

\begin{algorithm}[htb]
\caption{Cost scaling for warm-start $\varepsilon$-relaxation.}
\label{alg:cost-scaling}
\textbf{Input:} A graph $G=(V,E)$ and a dual prediction $\hat{p}$; The scaling parameter $c$.\\
\textbf{Output:} The optimal flow of $G$.
\begin{algorithmic}[1] 
\STATE Run preprocessing as in Algorithm~\ref{alg:warm-start};
\STATE Let $T$ be the maximal integer such that $c^T\le\|\hat{p}-p^*\|_{\infty}$.
\STATE Initialize the dual $p^{(T)}\leftarrow \lfloor(n+1)\hat{p}/c^T\rfloor$.
\FOR {each $t=T-1,\dots, 0$}
    \STATE Let $p^\prime\leftarrow cp^{(t+1)}$.
    \STATE Run $\varepsilon$-relaxation with the initial dual $p^\prime$, scaled edge costs $a^{(t)}=\lfloor (n+1)a/c^t\rfloor$, $\varepsilon=1$, and get dual solution $p^{(t)}$.
\ENDFOR
\STATE \textbf{return} the optimal flow for the dual $p^{(0)}$.
\end{algorithmic} 
\end{algorithm}
\subsection{Cost-scaling for warm-start $\varepsilon$-relaxation}
In practice, the vanilla version of the $\varepsilon$-relaxation algorithm is not scalable, since its time complexity grows linearly with respect to the cost scale $C$. Therefore, the vanilla algorithm is often combined with the $\varepsilon$-scaling trick to induce $O(n^3\log (nC))$ time complexity. This trick is also applicable to $\varepsilon$-relaxation with dual predictions. The algorithm is described in Algorithm~\ref{alg:cost-scaling}.

Note that $\varepsilon$-relaxation returns the optimal solution if $\varepsilon=\frac{1}{n+1}$. We can equivalently multiply the edge costs by $n+1$ and run the algorithm with $\varepsilon=1$. However, this does not reduce time complexity. In cost scaling, we solve a sequence of instances and gradually increase the scale of edge costs. Let $c$ be a scaling parameter (usually we choose $2\le c\le 4$). Initially we multiply edge costs by roughly $(n+1)/\|\hat{p}-p^*\|_{\infty}$ and run $\varepsilon$-relaxation. After the algorithm converges, we multiply edge costs and dual solutions by $c$, and repeat this process until edge costs are in the scale of $(n+1)a$. The algorithm is obviously correct and we analyze the time complexity in the following.

\begin{theorem}\label{thm:cost-scaling}
    The time complexity of Algorithm~\ref{alg:cost-scaling} is $O\left(\min\left\{n^3\log\|\hat{p}-p^*\|_{\infty},n^3\log(nC)\right\}\right)$. Specially, for 0/1 flow instances, this bound can be improved to $O\left(\min\left\{mn\log\|\hat{p}-p^*\|_{\infty},mn\log(nC)\right\}\right)$.
\end{theorem}
\begin{proof}
    Similar to Theorem~\ref{thm:time-complexity}, it suffices to prove the first terms of the minimum due to Lemma~\ref{lmm:preprocessing}. Since Algorithm~\ref{alg:cost-scaling} will call $\varepsilon$-relaxation at most $\log\|\hat{p}-p^*\|_\infty$ times, we only need to bound the time complexity of each call.
    
    Initially, $c^T=\Theta(\|\hat{p}-p^*\|_\infty)$. Therefore, the infinity norm prediction error of $p^{(T)}=\lfloor(n+1)\hat{p}/c^T\rfloor$ is $O(n)$. Since $\varepsilon=1$, the time complexity of the first scaling iteration $t=T-1$ is $O(n^3)$ (and $O(mn)$ for 0/1 flow) by Theorem~\ref{thm:time-complexity}.
    
    In subsequent iterations, for each $t=T-2,\dots,0$, we have $p^\prime=cp^{(t+1)}$ and $a^{(t)}-c\le ca^{(t+1)}\le a^{(t)}+c$. Since $p^{(t+1)}$ is the output of the last $\varepsilon$-relaxation step, $p^{(t+1)}$ satisfies 1-CS with respect to edge costs $a^{(t+1)}$. Therefore, we can simply verify that $p^\prime$ satisfies $2c$-CS with respect to $a^{(t)}$, which means for any path $H$, we have $d_H(p^\prime)\le 2c\cdot |H|=O(n)$, and thus $\beta(p^\prime)=O(n)$. The complexity of $\varepsilon$-relaxation for each step $t=T-2,\dots,0$ is hence $O(n^3)$ (and $O(mn)$ for 0/1 flow). Summing up the time for each step yields the desired bound.
\end{proof}

\begin{remark}\label{remark:tune}
    Algorithm~\ref{alg:cost-scaling} requires the knowledge of $\|\hat{p}-p^*\|_{\infty}$. Theoretically, we can eliminate this dependence by guessing the value of $T$ from 1 to $\lfloor\log(nC)\rfloor$. If calling $\varepsilon$-relaxation with $t=T-1$ does not return after $\Omega(n^3)$ time, we halt the algorithm and increase the value of $T$ by 1. Otherwise, we continue the algorithm to compute the result. This does not affect the time complexity of the algorithm. In practice, we can directly tune the value of $T$ empirically based on the learning error of the prediction.
\end{remark}

\section{Applications}\label{sec:app}
In this section, we introduce applications of the proposed algorithm, and discuss how to learn the dual prediction.

\subsection{Learning a fixed prediction}\label{sec:road}
Consider the problem of solving minimum-cost flow on a traffic network: Every day, an amount of goods is manufactured at specific locations and to be transported to several warehouses. The topology of the network, the capacity of edges, and the supplies of each node are fixed, but the edge costs can be seen as random variables, depending on the dynamic traffic conditions of every day. Although the minimum-cost flow may vary, the dual optimal solution may not change drastically. It will be time-consuming if we compute the optimal solution from scratch every day. Therefore, we can learn the expected optimal dual solution from collected data, and apply our warm-start $\varepsilon$-relaxation algorithm.

Let $\mathcal{D}$ be a distribution over minimum-cost flow instances, with a fixed network but randomly perturbed edge costs. We use $a\sim\mathcal{D}$ to denote a sample of edge costs from $\mathcal{D}$. Let $a^{(1)},a^{(2)},\dots,a^{(k)}\sim\mathcal{D}$ be $k$ identically independently distributed samples from $\mathcal{D}$. We are interested in learning a dual $\hat{p}$ based on the samples with low sample complexity.

Concretely, let $u(a,p):\mathbb{Z}^m\times\mathbb{R}^n\to [0,H]$ denote the running time performance of warm-start $\varepsilon$-relaxation for edge costs $a$ with the initial solution $p$, where $H$ is an upper bound of the performance\footnote{We assume $H$ is a constant. This assumption is standard in previous learning-augmented algorithms~\cite{dinitz21faster,chen22faster,davies23predictive}, and is indeed reasonable. We can always normalize the running time since graph topology is fixed.}. The generalization error of a learned dual $\hat{p}$ is defined as $\left|\mathbb{E}_{a\sim\mathcal{D}}[u(a,\hat{p})]-\frac{1}{k}\sum_{i=1}^k u(a^{(i)},\hat{p})\right|$. The sample complexity is the least number of samples such that the generalization error is at most $\varepsilon$ with high probability.

We have the following PAC-learning result, indicating we can learn a fixed prediction with sample complexity $\tilde{O}\left(n/\varepsilon^2\right)$. Since it is a uniform convergence bound, it is independent of the learning algorithm.
\begin{theorem}\label{thm:generalization}
    Let $a^{(1)},a^{(2)},\dots,a^{(k)}$ be $k$ i.i.d. samples from $\mathcal{D}$. If $$k=\Omega\left(\frac{H^2}{\varepsilon^2}\left(n\log(nC)+\log\left(\frac{1}{\delta}\right)\right)\right),$$ for any $\varepsilon>0$ and $\delta\in (0,1)$, with probability at least $1-\delta$, for any $p\in\mathbb{R}^n$, we have \begin{equation*}
        \left|\mathbb{E}_{a\sim\mathcal{D}}[u(a,\hat{p})]-\frac{1}{k}\sum_{i=1}^k u(a^{(i)},\hat{p})\right|\le\varepsilon.
    \end{equation*}
\end{theorem}
The proof technique of this theorem is to prove an upper bound on the pseudo-dimension of the function class of $u$. The proof is omitted to the appendix due to the page limit.
\begin{lemma}\label{lmm:pdim}
        Let $\mathcal{U}=\left\{u(\cdot,p)\,|\,p\in[0,(n-1)C]^n\right\}$, we have $\pdim(\mathcal{U})=O(n\log(nC))$.
    \end{lemma}
\begin{remark}
    Theorem~\ref{thm:generalization} is slightly different from sample complexity bounds in previous algorithm-with-prediction works~\cite{dinitz21faster,chen22faster,davies23predictive}. Previous works only consider the generalization error of the learned prediction, while we directly bound the error of the algorithm performance {\em end-to-end}, which is more general and practical. This follows the idea of {\em data-driven algorithm design} (see, e.g., \citet{balcan20data}).
\end{remark}
Now we discuss how to learn the prediction. Let $a^{(1)},a^{(2)},\dots,a^{(k)}$ be $k$ training samples, we first compute the optimal duals of these testcases. Since $u(a,\hat{p})=O(n^3\log\|\hat{p}-p^*\|_\infty)$, we can naturally learn the prediction by minimizing the surrogate loss, i.e., \begin{equation*}
    \min_{\hat{p}\in\mathbb{R}^n,\Delta\in\mathbb{R}^k} \frac{1}{k}\sum_{i=1}^k\log\|\hat{p}-p^{(i)}+\Delta_i\|_\infty,
\end{equation*} where $p^{(i)}$ is an optimal dual for $a^{(i)}$, $\Delta_i$ is the shift value by Fact 1. This problem can be converted into optimizing a convex function under linear constraints and solved by Frank-Wolfe:
\begin{align*}
    \min_{\hat{p},\Delta,M}\,&\frac{1}{k}\sum_{i=1}^k\log(M_i),\\
    \text{subj. to}\ & -M_i\le \hat{p}_j-p^{(i)}_j+\Delta_i\le M_i,\,\forall\,i,j,\\
    &M_i\ge 0,\,\forall\,i.
\end{align*}

\subsection{Learning a feature-based predictor}\label{sec:escape}
\begin{figure}
    \centering
    \includegraphics[width=0.85\linewidth]{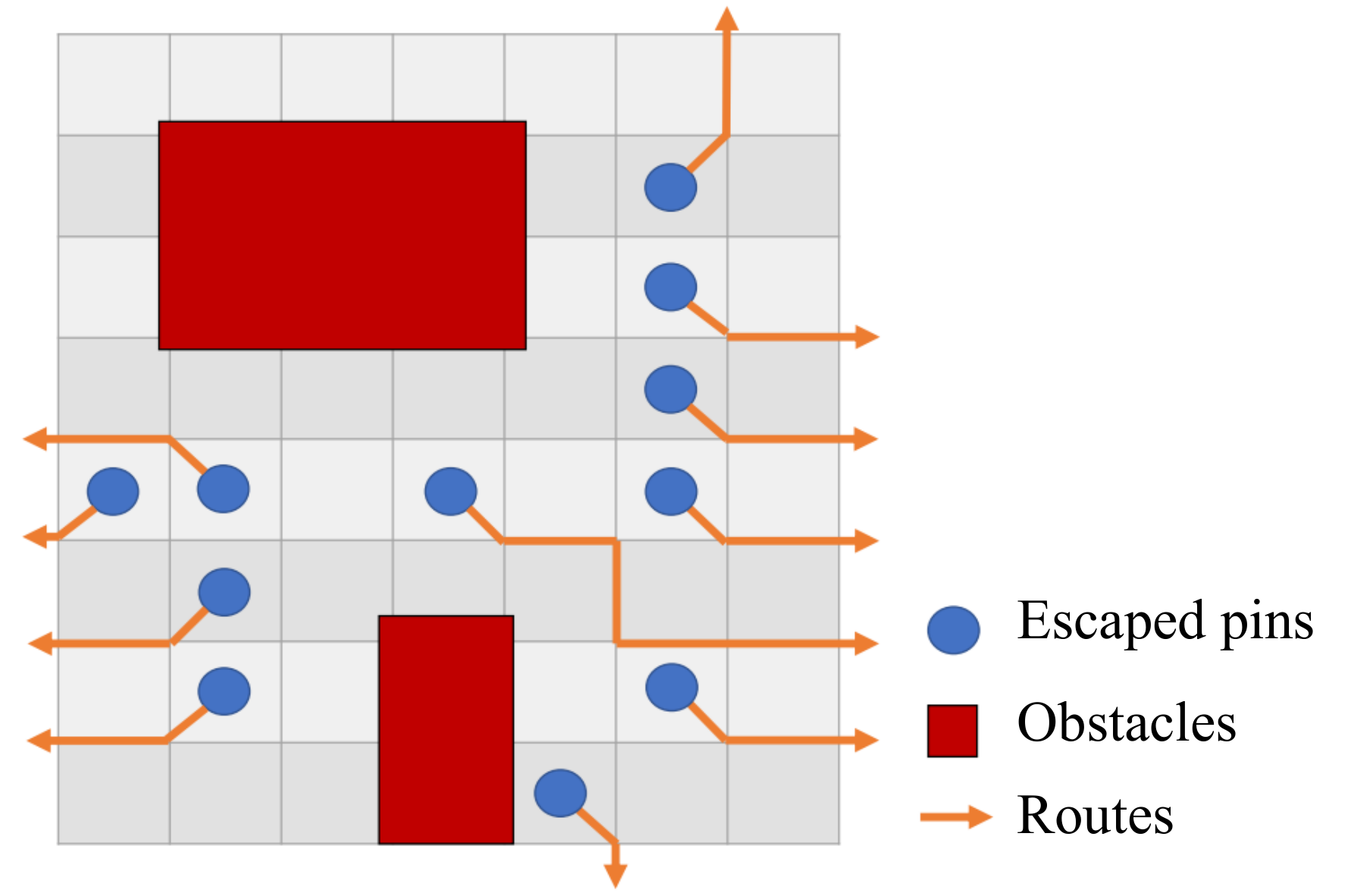}
    \caption{An example of unordered escape routing.}
    \label{fig:escape}
\end{figure}
A more general way of obtaining the prediction is to learn a mapping from a problem instance to the dual solution using machine learning models, such as neural networks. In this work, we apply this method to the escape routing problem in electronic design automation.

Escape routing is an important design automation problem in the design of printed circuit boards (PCB)~\cite{yan10recent}. In this work, we focus on the unordered escape problem: Given a $w\times h$ pin array with specific pins required to escape, find a routing solution that routes these pins to the boundary of the board without intersection. The total routed length should be minimized. Part of the routing grids may be occupied by obstacles and cannot be routed. See Figure~\ref{fig:escape} for an illustrating example.

Unordered escape can be modeled as a minimum-cost flow problem. Given the grid structure of the routing graph, we can naturally predict the dual solution using a convolutional neural network model. The routing grid is modeled as a $(w+1)\times (h+1)$ grid graph, and edges with unit capacity and geometric distance cost between adjacent nodes are added. Each pin is modeled as a supply node, with edges to its adjacent grid points. An artificial demand node is added to represent the destination with edges from boundary nodes. Each routing path is viewed as a unit flow from the pin to the boundary.

We introduce how to learn a feature-based neural network model for escape routing. The input to the model is a 3-channel $(w+1)\times (h+1)$ feature map. The 3 features include a binary value denoting whether it is occupied by obstacles, the shortest path distance to the nearest pin, and the shortest path distance to the boundary. The output is a $(w+1)\times (h+1)$ matrix representing the predicted dual. The dual of the destination is fixed to be 0, and the dual of pins is the maximum dual of adjacent grid nodes. To normalize the inputs and outputs, we divide the input features and output duals by $\max(w,h)$ so that the range of values is between $[0,1]$. We use a fully convolutional neural network with a UNet-style architecture~\cite{ronneberger15unet} as the prediction model. The model first transforms the input into 16-channel features using a $3\times 3$ convolution kernel. The backbone consists of two downsampling and two upsampling blocks, each of which block contains two $3\times 3$ convolution layers and two group normalization layers. Each pair of downsampling and upsampling layers with the same feature size is linked with residual connections. Finally, the model produces the output using a $3\times 3$ convolution kernel. We directly use the average mean-square loss (a.k.a., $l_2$ norm error) to train the neural network, since we find that $l_\infty$ norm error makes training very difficult.

We can similarly bound the sample complexity of the neural network model. Let $\mathcal{H}$ be a class of neural network functions that map the instance feature space to $[0,(n-1)C]^n$, the space of predicted duals. In other words, for a neural model $h\in\mathcal{H}$, it reads the feature of a minimum-cost flow instance and outputs an $\mathbb{R}^n$ vector, where the $i$-th component $h_i(\cdot)$ denote the initial dual of node $v_i$. We use $\mathcal{H}_i=\left\{h_i\ |\ h\in\mathcal{H}\right\}$ to denote the class of functions from $\mathcal{H}$ but limiting the output to be the $i$-th component. Let $d_{\mathsf{NN}}=\max_{i}\pdim(\mathcal{H}_i)$ denote the pseudo-dimension of the neural network architecture. Again, we use $u(\Phi,p)$ to denote the running time of the algorithm with dual $p$ on instance $\Phi$, and $\mathcal{D}$ is a distribution of $\Phi$.
\begin{theorem}\label{thm:nn}
    Let $\Phi^{(1)},\Phi^{(2)},\dots,\Phi^{(k)}$ be features\footnote{With a little abuse of notation, we use $\Phi$ to denote both a problem instance and its feature.} of $k$ i.i.d. samples from $\mathcal{D}$. If $$k=\Omega\left(\frac{H^2}{\varepsilon^2}\left(nd_{\mathsf{NN}}\log(nC)+\log\left(\frac{1}{\delta}\right)\right)\right),$$ for any $\varepsilon>0$ and $\delta\in (0,1)$, with probability at least $1-\delta$, for any $h\in\mathcal{H}$, we have \begin{equation*}
        \left|\mathbb{E}_{\Phi\sim\mathcal{D}}[u(\Phi,h(\Phi))]-\frac{1}{k}\sum_{i=1}^k u(\Phi^{(i)},h(\Phi^{(i)}))\right|\le\varepsilon.
    \end{equation*}
\end{theorem}
The proof of this theorem follows directly from Lemma~\ref{lmm:nn}.
\begin{lemma}\label{lmm:nn}
    Let $\mathcal{U}=\left\{u(\cdot,h(\cdot))\,|\,h\in\mathcal{H}\right\}$, $\pdim(\mathcal{U})=O(nd_{\mathsf{NN}}\log(nC))$.
\end{lemma}
\begin{remark}
    The pseudo-dimension of neural networks $d_{\mathsf{NN}}$ for typical neural architectures has been widely studied. For example, \citet{bartlett19nearly} gives an $O(WL\log W)$ upper bound for piecewise linear networks (e.g., networks with the ReLU activation) with $W$ weights and $L$ layers.
\end{remark}

\section{Experiments}\label{sec:exp}
We validate our theoretical results with empirical studies of two applications. We show that the $\varepsilon$-relaxation with predictions can significantly improve the running time, compared with the algorithm without predictions. We also implement two min-cost flow algorithms, network simplex (NS), and successive shortest paths (SSP). Experimental settings are omitted to the appendix.

\subsection{Synthesized predictions}
\begin{figure}
    \centering
    \includegraphics[width=0.5\linewidth]{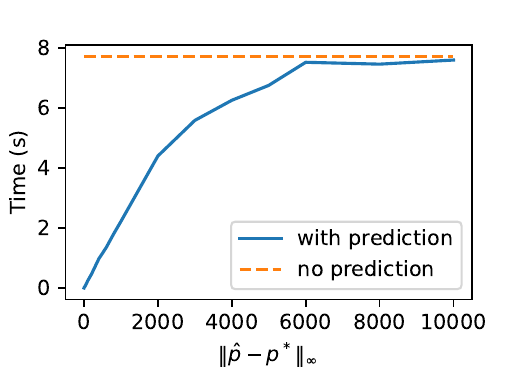}\includegraphics[width=0.5\linewidth]{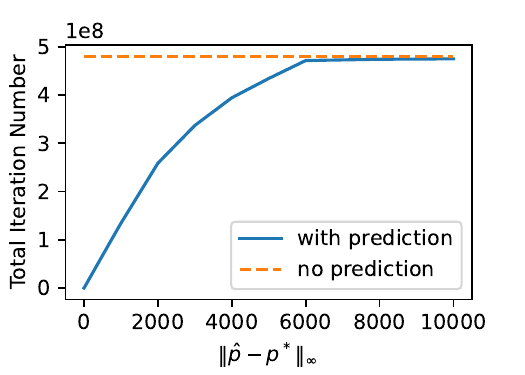}
    \caption{The running time and total number of iterations for the vanilla version of $\varepsilon$-relaxation on \texttt{paths\_01\_DC}.}
    \label{fig:vanilla}
\end{figure}

\begin{figure}
    \centering
    \includegraphics[width=0.5\linewidth]{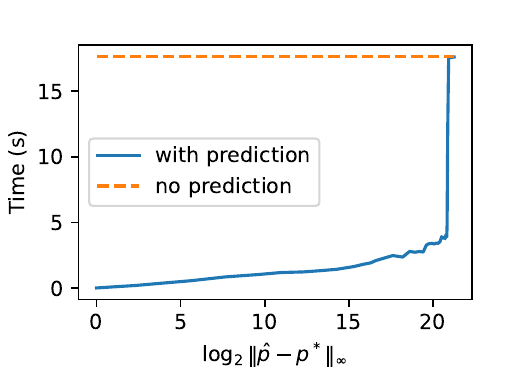}\includegraphics[width=0.5\linewidth]{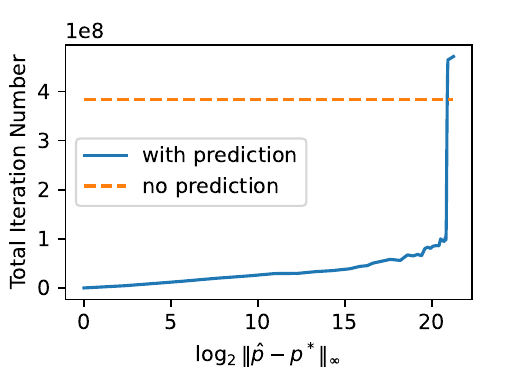}
    \caption{The running time and total number of iterations for the cost-scaling version of $\varepsilon$-relaxation on \texttt{flow\_03\_NH}.}
    \label{fig:scale}
\end{figure}

We first study the effect of the dual prediction by synthesized instances. Given an instance, we first compute the optimal solution, and randomly perturb the solution to synthesize predictions of different errors. Figures~\ref{fig:vanilla} and~\ref{fig:scale} illustrate the result of the vanilla and the cost-scaling versions of $\varepsilon$-relaxation. The efficiency of the algorithm grows with respect to the prediction error and is robust when the prediction is completely erroneous. The curves are not quite smooth due to the discreteness of rounding in Algorithm~\ref{alg:cost-scaling}. The growth of running time is steady when the error is small, but increases sharply when the error is sufficiently large. We explain the reason in the appendix.

\subsection{Road networks}
We study the application of traffic networks. For each instance, we sample 10 instances by randomly perturbing the edge costs, compute their optimal solutions, and learn a fixed prediction. Then, we test the algorithm with this prediction on the original instance. We use the cost-scaling version of $\varepsilon$-relaxation for comparison. Table~\ref{tab:road} illustrates the empirical results. We present results on large-scale instances with $>200000$ nodes. In the benchmark, there are 5 instances for each group. We report the average running time of each group. We achieve approximately $6.2\sim 21.4\times$ speedup using the learned prediction. Note that the groups of $\texttt{paths\_*}$ are 0/1 flow instances, $\varepsilon$-relaxation hence perform better, even outperforming network simplex without the help of predictions.

\subsection{Escape routing}
We also study the application of escape routing. We use the cost-scaling version of $\varepsilon$-relaxation for comparison. Table~\ref{tab:escape} illustrates the empirical results on 10 instances in the test set. We achieve approximately $1.1\sim 2.3\times$ speedup using the learned prediction. Note that the board size of the testcase varies from less than $300\times 300$ to more than $1000\times 1000$. We obtain the prediction using only a single neural network. This shows the generalization ability of the neural model across different problem sizes.
\begin{table}[]
    \centering\scriptsize
    \begin{tabular}{cccccc}\\\toprule
         Group & $n$  & $\varepsilon$-R (p.) & $\varepsilon$-R & NS & SSP  \\\midrule
         \texttt{paths\_04\_NV} & 261155 & \textbf{1.498}  & 24.247  & 19.293  & 11.684 \\
         \texttt{flow\_04\_NV} & 261155 & \textbf{18.214} & 112.926 & 20.223  & 51.211 \\
         \texttt{paths\_05\_WI} & 519157 & \textbf{3.875}  & 58.839  & 61.41   & 73.493 \\
         \texttt{flow\_05\_WI} & 519157 & \textbf{30.548} & 221.457 & 65.742  & 536.081 \\
         \texttt{paths\_06\_FL} & 1048506 & \textbf{5.193}  & 111.071 & 141.1   & 519.3 \\
         \texttt{flow\_06\_FL} & 1048506 & \textbf{45.946} & 630.73  & 271.133 & 1769.58 \\
         \texttt{paths\_07\_TX} & 2073870 & \textbf{25.866} & 323.846 & 694.118 & 2113.63 \\
         \texttt{flow\_07\_TX} & 2073870 & \textbf{238.41} & 2247.67 & 1127.91 & too long \\\midrule
         avg. & & \textbf{1.00} & 12.74 & 12.08 & 38.19 \\\bottomrule
    \end{tabular}
    \caption{Running times (sec) of min-cost flow algorithms on road network instances. $\varepsilon$-R denotes $\varepsilon$-relaxation, and (p.) denotes ``with prediction''.}
    \label{tab:road}
\end{table}
\begin{table}[]
    \centering\scriptsize
    \begin{tabular}{ccccccc}\\\toprule
        Case & $w\times h$ & \# pins & $\varepsilon$-R (p.) & $\varepsilon$-R & NS & SSP \\\midrule
        1 & $296\times 277$ & 710 & \textbf{2.103} & 3.302 & 11.749  & 11.546 \\
        2 & $366\times 418$ & 963 & \textbf{5.893} & 10.035 & 63.578  & 54.41 \\
        3 & $500\times 500$ & 1200 & \textbf{17.929} & 23.417 & 80.398  & 72.52 \\
        4 & $365\times 610$ & 1188 & \textbf{12.865} & 29.815 & 135.689 & 122.328 \\
        5 & $635\times 572$ & 1366 & \textbf{45.24} & 70.182 & 453.915 & 280.034 \\
        6 & $670\times 675$ & 1432 & \textbf{52.069} & 94.842 & 687.878 & 385.966 \\
        7 & $805\times 917$ & 1655 & \textbf{88.99} & 139.521 & 1572.02 & 1111.64 \\
        8 & $734\times 812$ & 2019 & \textbf{265.732} & 293.837 & 1943.52 & 1065.56 \\
        9 & $1003\times 965$ & 2001 & \textbf{126.141} & 245.585 & 2662.63 & 1616.98 \\
        10 & $1205\times 1374$ & 1985 & \textbf{630.487} & 955.19 & too long & 2592.41 \\\midrule
        avg. & & & \textbf{1.00} & 1.64 & 11.19 & 7.53 \\\bottomrule
    \end{tabular}
    \caption{Running times (sec) of min-cost flow algorithms on escape routing instances. ``\# pins'' denotes the number of pins to escape. Other abbreviations follow Table~\ref{tab:road}.}
    \label{tab:escape}
\end{table}

\section{Conclusion}
In this work, we show how to improve the $\varepsilon$-relaxation algorithm for min-cost flow with a machine-learned dual prediction. We prove strong time complexity bounds and study the sample complexity of learning the prediction. Experimental results illustrate considerable acceleration in two applications. For future work, we will explore other learning-augmented problems, e.g., general linear programs. We will also explore other neural models to improve the prediction accuracy to further accelerate the network flow algorithm. A question of interest is what kind of neural networks have stronger representation abilities for optimization.

\section{Acknowledgments}
This work is supported by the Major Research Plan of the National Natural Science Foundation of China (No. 92573107), and the Key Program of National Natural Science Foundation of China (No. 62034005).

\bibliography{bib}

\appendix

\section{Implementation Details of $\varepsilon$-Relaxation}\label{sec:algorithmdetail}
The concrete flow of $\varepsilon$-relaxation is listed in Algorithm~\ref{alg:eps-relaxation}. In the following, we introduce some implementation details of $\varepsilon$-relaxation.

\noindent\textbf{Cycling}. A special property of $\varepsilon$-relaxation is the so-called cycling phenomenon. We say $(v_i,v_j)$ is {\em unblocked} if $(v_i,v_j)\in E$, $x_{ij}<c_{ij}$ and $p_j-p_i-a_{ij}=\varepsilon$, (or conversely, $(v_j,v_i)\in E$, $x_{ij}>b_{ij}$ and $p_j-p_i-a_{ij}=-\varepsilon$), as the condition in line 10 of Algorithm~\ref{alg:eps-relaxation}. The unblocked edges form a directed graph, and cycling happens if this graph has cycles. The complexity analysis of $\varepsilon$-relaxation holds only when cycling does not happen. Theoretically, we can avoid cycling by following a few rules on the update order of nodes. See~\citet{bertsekas89parallel} for details. We omit these rules since it does not affect our analysis of $\varepsilon$-relaxation with predictions. In fact, we find that cycling does not happen in practical instances of minimum-cost flow, and hence we do not implement any cycle-avoiding rules in experiments. This slightly improves the practical efficiency of the algorithm.

\noindent\textbf{Heuristics}. Several heuristics can further improve the efficiency of $\varepsilon$-relaxation. For instance, we can perform surplus node iterations on both positive and negative surplus nodes. We can also adaptively set the value of $\varepsilon$ in the $\varepsilon$-scaling process. However, these heuristics do not have complexity guarantees. For fair comparisons, we do not implement these heuristics in experiments, but we do claim that our experimental results have the potential to be further optimized using these tricks.

\noindent\textbf{Parallelization}. The $\varepsilon$-relaxation algorithm is known to be suitable for parallel computation. In fact, we have the following result.
\begin{proposition}
    If the graph $G=(V,E)$ is $\Delta$-colorable, the time complexity of $\varepsilon$-relaxation both with and without predictions can be reduced to $\frac{\Delta}{n}T(n,m)$ using $O(n)$ cores in the shared memory model of parallel computation, where $T(n,m)$ is the serial time complexity.
\end{proposition}
\begin{proof}
    Suppose $V^\prime\subseteq V$ is a subset of nodes such that for any $u,v\in V^\prime$, $(u,v)\notin E$. It can be easily noticed that we can perform positive surplus node iteration (lines 10--14 of Algorithm~\ref{alg:eps-relaxation}) for nodes in $V^\prime$ in parallel without any conflicts. Therefore, if the graph $G=(V,E)$ is $\Delta$-colorable, we can do iterations for nodes with the same color simultaneously using $O(n)$ cores. The speedup ratio is at least $O(n/\Delta)$.
\end{proof}
The applications we considered in this work, traffic networks and escape routing, both have the potential for parallel acceleration. Since road networks in practice are usually planar graphs, they are 4-colorable. The routing grid graph of escape routing is obviously 3-colorable. However, for fair comparisons, we do not apply any parallel computing in experiments compared with other algorithms.

\begin{algorithm*}[htb]
\caption{The $\varepsilon$-relaxation algorithm.}
\label{alg:eps-relaxation}
\textbf{Input}: A graph $G=(V,E)$, $\varepsilon$.\\
\textbf{Output}: The optimal flow of $G$.
\begin{algorithmic}[1] 
\STATE Initialize an arbitrary dual solution $p$;
\STATE Initialize a (maybe infeasible) flow $x$ such that $(x,p)$ satisfies $\varepsilon$-CS;
\WHILE {$x$ is infeasible}
    \FOR {each node $v_i$ with $g_i>0$}
        \STATE Call $\textsc{Positive-surplus-node-iteration}(v_i)$;
    \ENDFOR
\ENDWHILE
\STATE Output the optimal flow $x$;
\STATE \textbf{function} $\textsc{Positive-surplus-node-iteration}(v_i)$
\STATE (Step 1:) Select $(v_i,v_j)\in E$ s.t. $x_{ij}<c_{ij}$ and $p_i-p_j-a_{ij}=\varepsilon$ and Goto Step 2; or select $(v_j,v_i)\in E$ s.t. $x_{ji}>b_{ji}$ and $p_j-p_i-a_{ji}=-\varepsilon$ and Goto Step 3;
\STATE (Step 2:) Let $\delta=\min\left\{g_i,c_{ij}-x_{ij}\right\}$, and set $$x_{ij}\leftarrow x_{ij}+\delta,\,g_i\leftarrow g_i-\delta,\,g_j\leftarrow g_j+\delta;$$ If $g_i=0$ and $x_{ij}<c_{ij}$, return; else Goto Step 1;
\STATE (Step 3:) Let $\delta=\min\left\{g_i,x_{ji}-b_{ji}\right\}$, and set $$x_{ji}\leftarrow x_{ji}-\delta,\,g_i\leftarrow g_i-\delta,\,g_j\leftarrow g_j+\delta;$$ If $g_i=0$ and $x_{ij}>c_{ij}$, return; else Goto Step 1;
\STATE (Step 4:) Set $p_i=\min_{\xi\in R^+\cup R^-}\xi$ where $$R^+=\left\{p_j+a_{ij}+\varepsilon\ |\ (v_i,v_j)\in E,\ x_{ij}<c_{ij}\right\},$$ $$R^-=\left\{p_j-a_{ji}+\varepsilon\ |\ (v_j,v_i)\in E,\ x_{ji}>b_{ji}\right\}.$$ Goto Step 1;
\STATE \textbf{end function}
\end{algorithmic} 
\end{algorithm*}

\section{Omitted Proofs}\label{sec:proof}
\subsection{Proof of Theorem~\ref{thm:generalization}}
\begin{proof}
    By Fact 1, we only consider $p\in [0,(n-1)C]^n$. To prove the generalization bound, we use the standard concept of pseudo-dimension in statistical learning theory.

    The pseudo-dimension is the natural extension of VC-dimension to real-valued functions, which characterizes the intrinsic complexity of a function class. Suppose $\mathcal{H}$ is a class of functions that map some domain $\mathcal{X}$ to $\left\{-1,1\right\}$. We say a finite set $S=\left\{x_1,\dots,x_m\right\}\subseteq \mathbb{X}$ is {\em shattered} by $\mathcal{H}$ if \begin{equation*}
        |\left\{\langle h(x_1),h(x_2),\dots,h(x_m)\rangle\ |\ h\in\mathcal{H}\right\}|=2^m.
    \end{equation*} The {\em VC-dimension} of $\mathcal{H}$ ($\vcdim(\mathcal{H})$) is the size of the largest set shattered by $\mathcal{H}$. Moreover, suppose $\mathcal{H}$ is a class of functions that map some domain $\mathcal{X}$ to $\mathcal{R}$. The {\em pseudo-dimension} of $\mathcal{H}$ ($\pdim(\mathcal{H})$) is the VC-dimension of $\mathcal{H}^\prime=\left\{h^\prime:(x,r)\mapsto\sgn(h(x)-r)\ |\ h\in\mathcal{H}\right\}$, where $r$ is called the {\em witness} of $h$. Classic learning theory gives the following uniform convergence bound via pseudo-dimension.
    \begin{proposition}[\citet{pollard84convergence}]\label{prop:uniform-convergence}
        Let $\mathcal{H}$ be a class of functions with domain $\mathbb{X}$ and range in $[0,H]$. For any distribution $\mathcal{D}$ over $\mathbb{X}$, any $\varepsilon>0$, and $\delta\in (0,1)$, if $$k=\Omega\left(\dfrac{H^2}{\varepsilon^2}\left(\pdim(\mathcal{H})+\log\left(\frac{1}{\delta}\right)\right)\right),$$ then with probability at least $1-\delta$ over $k$ i.i.d. samples $x_1,\dots,x_k\sim\mathcal{D}$, we have $$\left|\mathbb{E}_{x\sim\mathcal{D}}[h(x)]-\frac{1}{k}\sum_{i=1}^k h(x_i)\right|\le\varepsilon$$ for any $h\in\mathcal{H}$.
    \end{proposition}
    
    Equipped with pseudo-dimension, we can prove our sample complexity result by bounding the pseudo-dimension. Plugging Lemma 3 into Proposition~\ref{prop:uniform-convergence} yields the desired result.
\end{proof}
\subsection{Proof of Lemma~\ref{lmm:pdim}}
\begin{proof}
    Note that in Algorithm 2, the initial dual $\hat{p}$ is fed into the algorithm by setting $p^{(T)}=\lfloor (n+1)\hat{p}/c^T\rfloor$ in line 3. Due to the discreteness of $\lfloor\cdot\rfloor$, the function $u(a,\cdot)$ is a piecewise constant function. Concretely, since we have $$\lfloor (n+1)\hat{p}_i/c^T\rfloor=r\Longleftrightarrow \frac{c^Tr}{n+1}\le\hat{p}_i\le \frac{c^T(r+1)}{n+1}$$ for $1\le i\le n$, $T\ge 0$ and $r\in\left\{0,1,\dots,(n-1)C\right\}$, there exists a set of hyperplanes $$\left\{p_i=\alpha\ \bigg|\ p\in\mathbb{R}^n,i\in [1,n],\alpha\in\{0,\frac{1}{n+1},\dots,(n-1)C\}\right\}$$ that partitions $[0,(n-1)C]^n$ into $((n+1)(n-1)C+1)^n=(nC)^{O(n)}$ regions, such that $u(a,\cdot)$ is a constant function in each region.

     Suppose $\mathcal{U}$ shatters $k$ instances $a_1,\dots,a_k$. Since the parameter space $[0,(n-1)C]^n$ of $p$ is partitioned into $(nC)^{O(n)}$ constant-valued regions, and the partition is independent of $a$, the number of distinct tuples of $\langle u(a_1,\cdot),\dots,u(a_k,\cdot)\rangle$ is at most $(nC)^{O(n)}$. To shatter $k$ instances, it must be $2^k\le (nC)^{O(n)}$. Solving for the largest $k$ yields $k=O(n\log(nC))$. Therefore, $\pdim(\mathcal{U})=O(n\log(nC))$.
\end{proof}

\begin{table*}[]
    \caption{The running times (seconds) of each cost-scaling step on benchmark \texttt{paths\_04\_NV\_a} for $\varepsilon$-relaxation with and without the prediction.}
    \label{tab:breakdown}
    \centering\footnotesize
    \begin{tabular}{cccccccccccc}\\\toprule
         Cost-scaling step $t$ & 33 & 32 & 31 & 30 & 29 & 28 & 27 & 26 & 25 & 24 & $\le 23$ \\\midrule
         Without prediction & 2.96 & 3.04 & 2.64 & 1.70 & 2.43 & 2.17 & 2.49 & 1.08 & 1.02 & 0.68 & $<0.4$ \\
         With prediction & - & - & - & - & - & 1.20 & 1.07 & 0.94 & 0.71 & 0.61 & $<0.4$ \\\bottomrule
    \end{tabular}
\end{table*}
\subsection{Proof of Lemma~\ref{lmm:nn}}
\begin{proof}
    The following classic Sauer's lemma is required to prove our result.
    \begin{lemma}[\citet{mohri18foundation}]
         Let $\mathcal{H}$ be a function class from some domain $\mathbb{X}$ to $\left\{-1,1\right\}$ with $\vcdim(\mathcal{H})=d$. For any $S=\left\{x_1,\dots,x_k\right\}\subseteq\mathbb{X}$, we have $$\left|\left\{\langle h(x_1),\dots,h(x_k)\rangle\ |\ h\in\mathcal{H}\right\}\right|\le\sum_{i=0}^d\binom{k}{i}=O(k^d).$$
    \end{lemma}
    Let $\Phi^{(1)},\dots,\Phi^{(k)}$ be $k$ minimum-cost flow instance features. We first fix a node $v_t$ for some $1\le t\le n$. Let $\mathcal{H}^\prime_t=\left\{h_t:(\Phi,r)\mapsto\operatorname{sgn}(h_t(\Phi)-r)\,|\,h_t\in\mathcal{H}_t\right\}$. By definition, $\pdim(\mathcal{H}_t)=\vcdim(\mathcal{H}^\prime_t)$. We further define $$S^\prime=\left\{(\Phi_i,r)\ \bigg|\ i\in [1,k],r\in\{0,\frac{1}{n+1},\dots,(n-1)C\}\right\},$$ and thus $|S^\prime|=O(kn^2C)$. Applying Sauer's lemma yields the number of different tuples of $\langle\operatorname{sgn}(h_t(\Phi^{(i)}-r))\rangle_{(\Phi^{(i)},r)\in S^\prime}$ is $O((kn^2C)^{d_{\mathsf{NN}}})$.

    Now, we vary the fixed node $v_t$. By multiplying the number of choices for each $\mathcal{H}_t$, we have $$\left|\left\{\langle\operatorname{sgn}(h_t(\Phi^{(i)})-r_t)\rangle_{t,i,(x_i,r_t)}\ \Bigg|\ h\in\mathcal{H}\right\}\right|=O((kn^2C)^{nd_{\mathsf{NN}}}).$$ Recall that in the proof of Lemma 3, we claim that there exists a set of hyperplanes $$\left\{p_i=\alpha\ \bigg|\ p\in\mathbb{R}^n,1\le i\le n,\alpha\in\left\{0,\dots,(n-1)C\right\}\right\}$$ that partitions $[0,(n-1)C]^n$, such that $u(\Phi,\cdot)$ is a constant function in each region. Therefore, if the value of the tuple $\langle\operatorname{sgn}(h_t(\Phi^{(i)})-r_t)\rangle_{1\le t\le n,1\le i\le k,(x_i,r_t)\in S^\prime}$ is fixed, $u(\Phi,\cdot)$ is a constant. Therefore, as we vary the neural model $h\in\mathcal{H}$, there are at most $O((kn^2C)^{nd_{\mathsf{NN}}})$ values of $\langle u(\Phi^{(1)},h(\Phi^{(1)})),\dots,u(\Phi^{(k)},h(\Phi^{(k)}))\rangle$. To shatter $k$ instances, we have $2^k\le O((kn^2C)^{nd_{\mathsf{NN}}})$. Solving for the largest $k$ yields $k=O(nd_{\mathsf{NN}}\log(nC))$.
\end{proof}

\section{More Details of Experiments}\label{sec:detail}
\subsection{Experimental settings}\label{sec:setting}
\noindent\textbf{Environment.} All experiments are executed on a 2.0GHz Intel Core CPU. The network flow algorithms are implemented with \texttt{C++}, and the neural network is implemented with PyTorch. \texttt{O2} optimization is used for network flow algorithms.

\noindent\textbf{Benchmarks.} For traffic networks, we use TIGER/Line in the DIMACS challenge benchmark\footnote{Available at {http://lime.cs.elte.hu/\~{}kpeter/data/mcf/road/}.}, which is based on real-life road networks of the United States. The benchmark consists of two types of testcases. The \texttt{paths\_*} groups are 0/1 flow instances, and the \texttt{flow\_*} groups are normal instances. To simulate traffic conditions, we randomly perturb each edge cost independently with a Gaussian distribution. The distribution is zero-mean, with a standard deviation $\frac{1}{10}$ of the original edge cost.

For unordered escape routing, we collect 10 designs of PCBs and microfluidic chips with various sizes as the test set, and randomly generated 100 testcases of similar sizes as the training/validation set. After determining the board sizes and the number of pins, the pins and obstacles are uniformly sampled on the board. The percentage of obstacles is set to be 5\%.

\noindent\textbf{Baselines.} We compare the performances of $\varepsilon$-relaxation with and without predictions. We also implement two min-cost algorithms, network simplex (NS), and successive shortest paths (SSP). NS is widely used in practice due to its great empirical performance, and we follow the implementation of LEMON~\cite{lemon}, a widely used open-source solver. SSP is simple and easy to implement. Although its worst-case time complexity is worse than $\varepsilon$-relaxation, it may perform well on particular instances. For fair comparisons, we do \textbf{not} apply any multi-threaded computing, although we point out that $\varepsilon$-relaxation can be further accelerated using parallelization.

\noindent\textbf{Training details.} For escape routing, we train the neural network on 80 instances for 40 epochs and validate the model on 20 instances. We use the Adam optimizer, a batch size of 16, a learning rate $10^{-3}$, and divide the rate by 10 at epoch 15. After training, we use the neural model to infer the initial dual of $\varepsilon$-relaxation.

\noindent\textbf{Tuning cost-scaling parameter $T$.} In Remark 3, we claim that the cost-scaling parameter $T$ in Algorithm 2 is tunable in practice. In our experiments, we first compute the maximal integer $\hat{T}$ such that $c^{\hat{T}}$ is less than the empirical $l_\infty$ norm error. We find that setting the value of $T$ a little larger can empirically improve algorithm performance. Therefore, in our experiments, we set $T=\hat{T}+2$.

\subsection{Training loss for the neural model}
\begin{figure}
    \centering
    \includegraphics[width=0.7\linewidth]{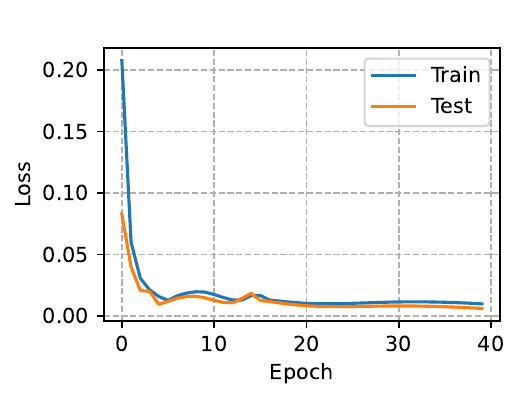}
    \caption{The training curve of the neural network prediction model.}
    \label{fig:curve}
\end{figure}

The training curve of the neural network model for our escape routing application is shown in Figure~\ref{fig:curve}. Note that the loss is normalized by $\min(w,h)$ to make it lie between $[0,1]$.

\subsection{Running time breakdown}\label{sec:breakdown}
We study the empirical improvements of the running time induced by the prediction. The key finding is that the improvement is often larger than theory for most practical instances. For example, Table~\ref{tab:breakdown} lists the running time breakdown of each cost-scaling step for $\varepsilon$-relaxation with and without predictions. It is easy to notice that the running time of each step gradually decreases. Although the time complexity is $O(n^3)$ theoretically for each step, the empirical running time is often smaller for small $t$. This explains the sharp increase in Figure 3, and shows that the theoretical analysis is pessimistic for network flow problems in practice. Obtaining tighter complexity bounds is a potential direction for future research.

\end{document}